\documentclass{svproc}

\usepackage{type1cm}        
\usepackage{makeidx}         
\usepackage{graphicx}        
\usepackage{multicol}        
\usepackage[bottom]{footmisc}
\usepackage[tight,footnotesize]{subfigure}

\usepackage{newtxtext}       %
\usepackage[varvw]{newtxmath}       

\makeindex             

\begin{document}

\title{MARLAS: Multi Agent Reinforcement Learning for cooperated Adaptive Sampling}
\def\thefootnote{*}\footnotetext{Equal contribution.}\def\thefootnote{\arabic{footnote}}
\author{Lishuo Pan$^{*1,2}$, Sandeep Manjanna$^{*1,3}$, and M. Ani Hsieh$^{1}$}
\institute{$^{1}$GRASP Laboratory at the University of Pennsylvania, Philadelphia, USA,\\
$^{2}$Brown University, Providence, USA,\\
$^{3}$Plaksha University, Mohali, India,\\
\email{lishuo\_pan@brown.edu}\\
\email{msandeep.sjce@gmail.com}\\
\email{m.hsieh@seas.upenn.edu}}
%
%
\maketitle


\begin{abstract}
The multi-robot adaptive sampling problem aims at finding trajectories for a team of robots to efficiently sample the phenomenon of interest within a given endurance budget of the robots. In this paper, we propose a robust and scalable approach using Multi-Agent Reinforcement Learning for cooperated Adaptive Sampling (MARLAS) of quasi-static environmental processes. Given a prior on the field being sampled, the proposed method learns decentralized policies for a team of robots to sample high-utility regions within a fixed budget. The multi-robot adaptive sampling problem requires the robots to coordinate with each other to avoid overlapping sampling trajectories. Therefore, we encode the estimates of neighbor positions and intermittent communication between robots into the learning process. We evaluated MARLAS over multiple performance metrics and found it to outperform other baseline multi-robot sampling techniques. Additionally, we demonstrate scalability with both the size of the robot team and the size of the region being sampled. We further demonstrate robustness to communication failures and robot failures. The experimental evaluations are conducted both in simulations on real data and in real robot experiments on demo environmental setup\footnote{The demo video can be accessed at: \url{https://youtu.be/qRRpNC60KL4}}.
\end{abstract}



\vspace{-2.0em}
\section{Introduction}
\label{sec:introduction}
\vspace{-1.0em}

\begin{figure}
    \centering
    \includegraphics[width=0.99\textwidth]{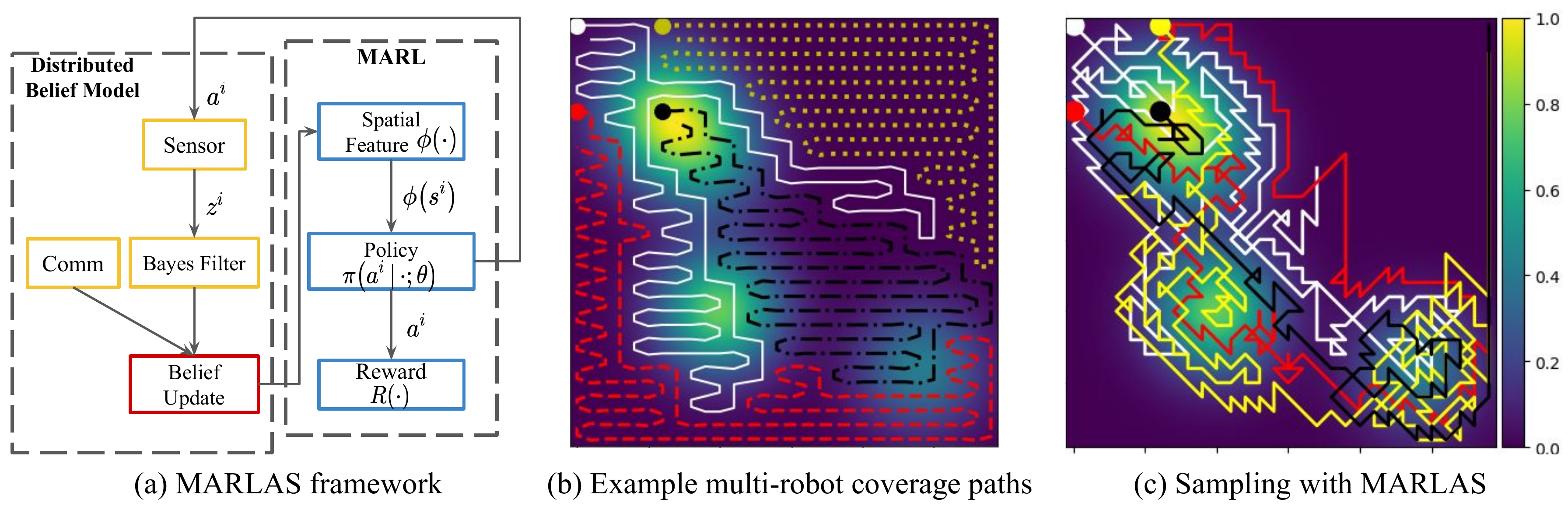}
    \caption{(a) Overview of MARLAS algorithm, (b) coverage paths generated by one of the area division approaches (DARP~\cite{kapoutsis2017darp}), and (c) an example sampling run of the proposed approach (MARLAS).}
    \vspace{-2em}
    \label{fig:framework}
\end{figure}

In this paper, we propose a decentralized multi-robot planning strategy (Fig.~\ref{fig:framework}(a) presents the framework of our method) to cooperatively sample data from two-dimensional spatial fields so that the underlying physical phenomenon can be modeled accurately. 
We consider sampling from quasi-static spatial processes, that is, we can reasonably assume the field to be temporally static for the sampling duration. Examples of such spatial fields include algal blooms, coral reefs, the distribution of flora and fauna, and aerosol concentration. Building high-resolution representations of environmental physical phenomena can help better understand the effects of global warming and climate change on our environment. 
Observing and modeling the spatial evolution of such fields using multiple data sampling robots plays a key role in applications such as environmental monitoring, search and rescue, anomaly detection, and geological surveys. We present a planning algorithm for multiple robotic platforms to sample these spatial fields efficiently and adaptively. In this context, sampling refers to collecting sensor measurements of a phenomenon.


Adaptive sampling of spatial fields with hotspots has gained a lot of momentum in the field of robotic exploration and mapping~\cite{low2008adaptive,sadat2015fractal,manjanna2018policy,almadhoun2019survey}. 
Adaptive sampling refers to strategic online path planning for robots based on the observations made until the current time step. Exhaustively sampling each point of an unknown survey region can be tedious, inefficient, and impractical if the survey space is large and/or the phenomenon of interest has only a few regions (\emph{hotspots}) with important information~\cite{low2008adaptive,salam2019adaptive}. It has been observed that for low-pass multiband signals, uniform sampling can be inefficient, and sampling rates far below the Nyquist rate can still preserve information~\cite{venkataramani2000perfect}. This is the key guiding principle behind active and non-uniform sampling~\cite{rahimi2005adaptive}. Although the approaches solving coverage control problems~\cite{cortes2004coverage,durham2011discrete} seem suitable for solving sampling problems, they differ in a quite important aspect. The coverage control or area division problem~\cite{aurenhammer1991voronoi,breitenmoser2010voronoi} focuses on finding the most preferable positions for the robots such that a non-occupied space is covered with their sensors or the corresponding assigned areas are physically visited by the respective robot as depicted in Fig.~\ref{fig:framework}(b). On the contrary, in this paper we are looking at designing robot paths that visit high-utility regions within a fixed budget as illustrated in an example run of our method in Fig.~\ref{fig:framework}(c).

Recently, multi-agent reinforcement learning (MARL) methods have proven to be effective for searching cooperative policies in complex and large-scale multi-robot tasks and they have been explored for information gathering and coverage problems. Persistent monitoring and coverage with MARL involve defining equal utility for every state~\cite{mishra2021galopp,chen2021multi}. In the sampling problem, utilities are distributed unevenly, thus requiring nonuniform exploration. Some of the multi-robot information gathering techniques are limited to offline planning~\cite{kantaros2021sampling} or are incapable of handling robot failures, as they function by dividing the region of interest among the robots in the team~\cite{kapoutsis2017darp}. Scalable multi-agent reinforcement learning policy is proposed to achieve large-scale active information gathering problem in~\cite{hsu2021scalable}. However, they use a central node for target localization, which is fragile to single-node failure. 

The key contribution of this work is a fully decentralized multi-robot control strategy, MARLAS, to learn multi-robot cooperation and achieve efficient online sampling of an environmental field. From comprehensive experimental results, we show that our method is scalable with both the size of robot team and the size of the region being sampled, and achieves equitable task distribution among robots. We further demonstrate our approach is robust to both communication failures and individual robot failures. 


\section{Problem Formulation}
\label{sec:problem_formulation}

Consider the multi-robot adaptive sampling problem where $N$ homogeneous robots aim to maximize accumulated utilities over a region of interest in a given time horizon $H$. The sampling region $\mathcal{E} \in \mathbb{R}^2$ is a bounded continuous $2D$ space. In this work, $\mathcal{E}$ is discretized into uniform grid cells forming a workspace with length $h$ and width $w$. Thus, the position $\mathbf{x}^{i}$ of robot $r^{i}$ is an element of $\mathbb{Z}^{2}$. Each grid is assigned a prior value $F(x_{1},x_{2})$, which represents the amount of data that robots can collect at this location. In this work, we assume once the grid is visited by any robot, all the data in that grid will be collected. We further assume that a low-quality estimate of reward map $F \in \mathbb{R}^{h\times w}$ of the region $\mathcal{E}$ can be obtained through pilot surveys or satellite data or historical knowledge of the field in the region of interest.

The objective is to obtain a set of decentralized policies $\boldsymbol\pi$ for all robots in the multi-robot team, that maximize the averaged expected return of $N$ robots $J_{tot}(\boldsymbol\pi) = \frac{1}{N} \sum_{i=1}^{N} J^{i}(\pi^{i})$. Here, $J^{i}$ is the expected return of the robot $r^{i}$ along its trajectory $\tau^{i}$. 
In the homogeneous team setting, we can assume that all robots have the same control strategy; therefore, we will drop the subscripts and write the policy as $\pi = \pi^{i}$ for each robot $r^{i}$. We use a neural network with parameters $\boldsymbol\theta$ to approximate the underlying optimal control policy as a conditional probability $\pi_{\boldsymbol\theta} = P(\mathbf{a}^{i}|\mathbf{s}^{i};\boldsymbol\theta)$, where $\mathbf{a}^{i}$, $\mathbf{s}^{i}$ are the action and state of the robot $r^{i}$, respectively. 
Each robot maintains locally a belief of the position of other robots $\hat{\mathbf{x}}^{i}_{j} \in \mathbb{R}^{h\times w}$ and the belief of the reward map $\hat{F}^{i} \in \mathbb{R}^{h\times w}$ throughout the mission. Our learning task is to solve the optimization problem to find a set of parameters $\boldsymbol\theta^{*}$ for the neural network.

\textbf{Multi-robot Learning Formulation}\label{subsec:formulation_MAL} In our decentralized multi-agent system setting, individual agents do not have the direct access to the ground truth information, and need to trade-off between exploiting the current information and communicating with neighbors to solve the task in a cooperative manner. We frame the task as a decentralized partially observable Markov Decision Process (Dec-POMDP) represented by the tuple, $(\mathcal{I}, \mathcal{S}, \{\mathcal{A}^{i}\}, \{\mathcal{Z}^{i}\}, \mathcal{T}, R, \gamma)$. Here, $\mathcal{I}$ is the set of robots, $\mathcal{S}$ is a set of states, and $\mathcal{A}$ is the set of discrete actions. 
$\mathcal{Z}^{i}$ is a set of possible observations by the robot $r_{i}$. $\mathcal{T}$ models the transition function. $R$, $\gamma$ are the reward function and discount factor. Optimal solution of the Dec-POMDP using dynamic programming is provably intractable as the time horizon, number of robots, and the size of region increases~\cite{dibangoye2016optimally}. 
In this work, we train a neural network in a multi-agent reinforcement learning (MARL) setting to approximate the underlying optimal decentralized policy. 

\section{Approach}
\label{sec:approach}
An overview of our proposed MARLAS framework is presented in Fig.~\ref{fig:framework}(a). Let $\{r^{1},r^{2},\cdots, r^{N}\}$ be the set of robots, and $\left(\mathbf{x}^{i}_{(t-l+1):t}, \{\mathbf{x}^{j}_{(t-l+1):t}\}, F_{(t-l+1):t}\right)$ represents the state of each robot, where $\mathbf{x}^{i}_{(t-l+1):t}$ are robot's historical positions up to $l$ steps in the past, $\{\mathbf{x}^{j}_{(t-l+1):t} , \forall j \neq i \}$ is the set of positions of the neighboring robots with a history up to $l$ steps in the past, and 
$F_{(t-l+1):t}$ are the reward maps up to $l$ steps in the past. Thus, in our formulation each robot stores the trajectory history of length $l$ and uses intermittent communication with other robots to update the neighbor positions and reward maps. The individual robot has no direct access to the neighbor positions when they are not in communication. Thus, the ground truth of the reward maps quickly become unavailable to individual robots. As a substitute, each robot $r^{i}$ estimates and maintains belief distributions over the neighbor positions and the reward maps. Hence, the new state of the robot $r^{i}$ at time $t$ is defined as $\mathbf{s}^{i}_{t}:= \left(\mathbf{x}^{i}_{(t-l+1):t}, \{\hat{\mathbf{x}}^{i}_{j,(t-l+1):t}\}, \hat{F}^{i}_{(t-l+1):t} \right)$. We denote $\hat{\mathbf{x}}^{i}_{j,(t-l+1):t}$ as belief distributions of neighbor positions and $\hat{F}^{i}_{(t-l+1):t}$ as the belief distributions of reward maps.
A discrete action set is defined as $\mathcal{A}=\left\{(i,j) \mid \forall i,j \in \{-1,0,1\}, (i,j)\neq (0,0) \right\}$, that is our action space consists of actions leading to $8$-connected neighboring grids from the current position. At each time step $t$, each robot $r^{i}$ takes an action $\mathbf{a}^{i}_{t}\in \mathcal{A}$. We assume that robots can uniquely identify other robots, and thus ${\mathbf{z}^{i}_{j,t}}$ represents the observation of robot $r^{j}$'s position by robot $r^{i}$ at time $t$. More concretely, we define ${\mathbf{z}^{i}_{j,t}}$ as the occupancy $ \{0,1\}^{h\times w}$ of grid cells within the sensing radius of robot $r^{i}$. We assume a deterministic transition function $\mathcal{T}$, but our method can be easily extended to handle non-deterministic transitions.

\textbf{Distributed Belief State Estimate}
\label{subsec:state_estimate}
We assume that a robot's estimates of its neighbor positions is limited by its sensing radius and these estimates are noisy due to hardware limitations and environmental uncertainties. Direct communication between robots is limited by the communication radius. Since knowledge of global robot states is necessary to achieve multi-robot coordination, we assume each robot $r^{i}$ maintains local beliefs $\hat{\mathbf{x}}^{i}_{j}$ of all teammates' positions, where $j$ is the index of its teammates. Without loss of generality, we choose the Bayes filter to estimate neighbor positions, which can be replaced with other filters such as the Kalman or learning-based filters.


For simplicity, we assume that the sensing radius is equal to the communication radius, and both denoted as $d_{cr}$, but this is not a strict requirement. When $\mathbf{z}^{i}_{j}(x_{1}, x_{2})=1$, a neighboring robot is detected at location $(x_{1}, x_{2})$, otherwise $\mathbf{z}^{i}_{j}(x_{1}, x_{2})=0$. Due to sensing noise, we consider false positive and false negative in the discrete setting. Grid cells that are outside of the robot's sensing radius have $\mathbf{z}^{i}_{j}=0$. 
In this work, we assume that the robot has no knowledge of its neighbors' control policies. In other words, in the action update of a neighbor's position estimate, the unobserved neighbors have equal transition probabilities to its $8$-connected neighboring grids. 
Thus, the distribution $\hat{\mathbf{x}}^{i}_{j}$ is dictated by a $2D$ random walk in the action update. The sequence of action and sensor updates are given by,
\begin{eqnarray}
\textrm{action update:} & \quad p(\hat{\mathbf{x}}^{i}_{j,t+1}|\hat{\mathbf{x}}^{i}_{j,t}, \mathbf{a}^{j}_{t}) = p(\hat{\mathbf{x}}^{i}_{j,t+1}|\hat{\mathbf{x}}^{i}_{j,t}),\\
\textrm{sensor update:} & \quad
p(\hat{\mathbf{x}}^{i}_{j,t+1}|\mathbf{z}^{i}_{j,t}) = \frac{p(\mathbf{z}^{i}_{j,t}|\hat{\mathbf{x}}^{i}_{j,t})\cdot p(\hat{\mathbf{x}}^{i}_{j,t})}{\sum_{\hat{\mathbf{x}}^{i}_{j,t}} p(\mathbf{z}^{i}_{j,t}|\hat{\mathbf{x}}^{i}_{j,t})\cdot p(\hat{\mathbf{x}}^{i}_{j,t})}
\end{eqnarray}

To enhance each robot's estimates of its neighbors' positions, we utilize communication between robots. To achieve this, each robot saves a length $l$ of the trajectory history $\tau^{i}_{t-l+1:t}$, considering the limitation of memory units in distributed systems. When neighboring robots enter a robot's  communication radius, the robot has direct access to its neighbor's trajectory history and updates its belief of the neighbor's position, $\hat{\mathbf{x}}^{i}_{j,t-l+1:t}$, accordingly. In this work, we only consider $1$-hop communication without communication delays. In general, multi-hop communication would introduce communication delays and thus is out of scope of this work. 

\textbf{Decentralized Control Policy}
In this work, we use the 
expressiveness of a neural network to approximate the underlying optimal decentralized control policy for the proposed multi-agent reinforcement learning problem. We use a fully connected neural network of two hidden layers $f(\cdot)$ with $128$ hidden units, and hyperbolic tangent activation functions following each hidden layer. The output layer units correspond with actions to be taken. To define a stochastic policy, a softmax layer is attached to generate a distribution on the set of actions $\mathcal{A}$. From experiments, we find that with the spatial feature aggregation $\phi(\cdot)$ detailed in the following section, two-hidden-layers neural networks have sufficient expressiveness to approximate the underlying optimal policy. We do not exploit the fine-tuning of neural network architectures, as it is not the main focus of this work. Policy parameterization is given by
\begin{equation}
    \pi_{\boldsymbol\theta}(\mathbf{a}^{i}_{t}|\mathbf{s}^{i}_{t}) = \text{Softmax}(f(\phi(\mathbf{s}^{i}_{t}); \boldsymbol\theta))\label{eq:NN_policy}
\end{equation}
where $\phi(\mathbf{s}^{i}_{t})$ represents the spatial feature aggregation vector. At each time step, robots sample the actions from stochastic policy and move to their next positions following transition function $\mathcal{T}$.

To avoid collisions with other robots in their neighborhood, robots employ a potential field based obstacle avoidance strategy similar to~\cite{khatib1986real}. 
It is assumed that collision avoidance will only be activated within $d_{0}$ distance from the closest neighboring robot.  
We further quantized repulsive force synthesised from collision avoidance strategy, by maximizing its alignment with the action from the discrete set $\mathcal{A}$. For robots whose collision avoidance is active, their actions generated from learned policy are replaced with a repulsive force. 
Additionally, robots generate random actions when the collision occurs.

\textbf{Spatial Feature Aggregation}
\label{subsec:feature_design}
In our previous work~\cite{manjanna2018policy,manjanna2021scalable}, we proposed and evaluated various feature aggregation methods for adaptive sampling. Multiresolution feature aggregation~\cite{manjanna2018policy} has been empirically proven to outperform other methods, as it reduces the state dimension and utilizes the geometry induced bias for the sampling task. In multiresolution feature aggregation, we divided the current belief distribution of reward map $\hat{F}^{i}_{t}$ into regions with different resolutions according to their distance to each robot. Feature regions closer to a robot have higher resolution, and resolution decreases farther away from the robot. Since rewards at locations visited by other robots are removed, each robot must estimate its neighbors' positions at any given time. We utilize each robot's belief estimates for other robot locations, $\hat{\mathbf{x}}^{i}_{j}$ and the belief estimate for previous reward map $\hat{F}^{i}_{t-l}$ when updating the belief of current reward map $\hat{F}^{i}_{t}$ in the form 
$\hat{F}^{i}_{t}(x_{1},x_{2}) = \hat{F}^{i}_{t-l}(x_{1},x_{2}) \left[\mathcal{J}_{h,w}-\sum_{k=t-l+1}^{t}\sum_{j}\hat{\mathbf{x}}^{i}_{j,k}(x_{1},x_{2})\right]_{+}$, for all $(x_{1},x_{2})$. 
Here, $\mathcal{J}_{h,w}$ is a all-ones matrix of size of the reward map, the $\hat{\mathbf{x}}^{i}_{j,k}$ is the discrete belief distribution of the neighbor positions with its value between $0$ and $1$ in each grid cell, the operation $\left[\cdot\right]_{+}$ converts the negative value to $0$. Notice that the direct communication between robots updates the $\hat{\mathbf{x}}^{i}_{j,t-l+1:t}$, thus we recompute the summation over $l$ time steps of the belief distributions of the neighbor positions. Here, the belief distributions of reward maps $\hat{F}^{i}_{(t-l+1):t}$ follow the Markov property.

\textbf{Multi-robot Learning in Adaptive Sampling}
\label{subsec:MARL}
As mentioned in Section~\ref{subsec:formulation_MAL}, solving a Dec-POMDP optimally is provably intractable in our task. Therefore, we use the belief model to maintain belief distributions of the ground truth and employ a neural network to approximate the optimal decentralized policy $\pi^{*}$ and we use policy gradient methods to search the optimal policy on the neural network parameter space. 
To encourage cooperation, we adapt the centralized training with decentralized execution (CTDE) technique commonly used for multi-agent policy training. More specifically, we compute a centralized reward as the average of all the robots' individual rewards, using global reward function $R(\cdot)$, state $\mathbf{s}$ and action $\mathbf{a}$ information. Thus, the centralized expected return is given by $J_{tot}(\pi_{\boldsymbol\theta}) = \mathbb{E}_{\tau\sim \pi_{\boldsymbol\theta}}\left[\frac{1}{N}\sum_{t=1}^{H}\gamma^{t}\sum_{i=1}^{N}R\left(\mathbf{s}^{i}_{t}, \mathbf{a}^{i}_{t}\right)\;\middle|\; \mathbf{s}_{0} \right],$
Control policy $\pi_{\boldsymbol\theta}$ is decentralized in both training and execution. 

To optimize $\boldsymbol\theta$ directly, we use policy gradient
which gives the gradient of the expected return in the form of
$\nabla J(\pi_{\boldsymbol\theta}) \propto \mathbb{E}_{\tau\sim\pi_{\boldsymbol\theta}}\left[G_{t}\cdot \nabla \log \pi_{\boldsymbol\theta}(\mathbf{a}_{t}|\mathbf{s}_{t}) \right]$, where $G_{t} = \sum_{k=t+1}^{T}\gamma^{k-t-1}R\left(\mathbf{s}^{i}_{k}, \mathbf{a}^{i}_{k}\right)$. Action $\mathbf{a}_{t}$ and state $\mathbf{s}_{t}$ are drawn from $\tau$. 
To estimate $G_{t}$, we adapt the Monte Carlo based REINFORCE method by generating trajectories $\tau$ following $\pi_{\boldsymbol\theta}$. 
For each training episode of MARLAS, we initialize robots randomly in ${\cal E}$. We assumed the robot knows the initial position of all the other robots. Each robot is given the control policy with the most updated parameters $\boldsymbol\theta$. At time step $t$, each robot $r^{i}$ randomly samples action $\mathbf{a}^{i}_{t}$ from $\pi_{\boldsymbol\theta}(\mathbf{a}^{i}_{t}|\mathbf{s}^{i}_{t})$. Robot $r^{i}$ receives a reward $F(\mathbf{x}^{i})$ by sampling data at location $\mathbf{x}^{i}$. Once the grid cell $\mathbf{x}^{i}$ is visited, $F(\mathbf{x}^{i})$ is set to a negative value indicating the data has been consumed at this location. If multiple robots enter the same grid cell simultaneously, the reward will be evenly distributed to each robot in the cell, and there will be a collision penalty for each robot in the same cell. The reward function is given by,
\begin{equation}
    R\left(\mathbf{s}^{i}, \mathbf{a}^{i}\right) = \frac{F(\mathbf{x}^{i})}{c} + \beta_{col}\cdot \mathbb{I}(\mathbf{x}^{i}=\mathbf{x}^{j}), ~\forall j \in \mathcal{I}\label{eq:reward},
\end{equation}
where $c$ is the total number of robots in the grid of $\mathbf{x}^{i}$, $\mathbb{I}$ is the indicator function, which indicates a collision between robots, and $\beta_{col} < 0$ is the collision penalty.  Each robot generates $\tau$ by executing the control policy up to the time horizon $H$. Furthermore, we generate $M$ trajectories for each robot as one training batch. We use REINFORCE with baseline to update the parameters $\boldsymbol\theta$ as follows
\begin{equation}
    \boldsymbol\theta^{k+1} = \boldsymbol\theta^{k} + \alpha\cdot  \frac{1}{N\cdot M}\sum_{i=1}^{N} \sum_{m=1}^{M} \sum_{t=0}^{H-1} (G^{i,m}_{t}-b_{t}^{i})\nabla \log \pi_{\boldsymbol\theta}(\mathbf{a}^{i,m}_{t}|\mathbf{s}^{i,m}_{t})~,
\end{equation}
with the baseline defined as $b_{t}^{i}=\frac{1}{M} \sum_{m=1}^{M} G_{t}^{i,m}$. Superscription $i,m$ are the indices of the robots and the trajectories. Note that we draw actions according to the policy distribution in training to enhance the exploration. During deployment, we draw the action that has the maximum probability.

%

%
\section{Experiments and Results}
\label{sec:results}

We train the multi-robot team on a region $\mathcal{E}$ synthesized by a mixture of Gaussians. 
In the training phase, we use a robot team size of $N=5$, a sensing and communication radius of $d_{cr}=10$ grid cells, a collision avoidance range of $d_{o}=2$ grid cells and collision penalty coefficient $\beta_{col}=-2$. Other hyperparameters used for training include: a discount factor $\gamma=0.9$, training time horizon $H=200$, and set the number of trajectories $M=40$. We use the ADAM stochastic optimizer\cite{kingma2014adam} in our training phase and the control policy is trained on a single map of size $30\times30$ by changing the starting locations of the robots for every epoch. We only use one trained policy to generate all the results in this paper. We found that the learned policies can adapt to diverse fields during the test phase as presented in our results. The results presented in this Section show that the learned control policies outperform existing baseline methods. We further evaluate scalability of MARLAS for different robot team and workspace sizes, and the robustness of MARLAS to communication failures and robot failures.

\begin{figure}
    \centering
    \subfigure[Bathymetry data]{\includegraphics[width=0.26\textwidth]{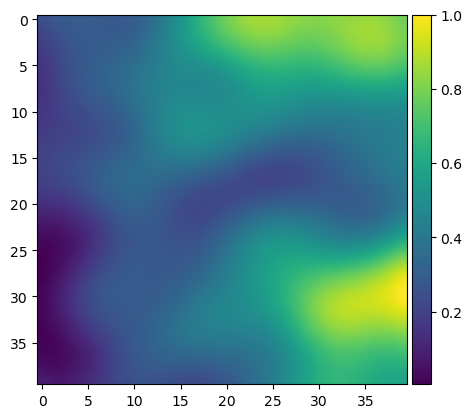}} 
    \subfigure[Synthetic data]{\includegraphics[width=0.233\textwidth]{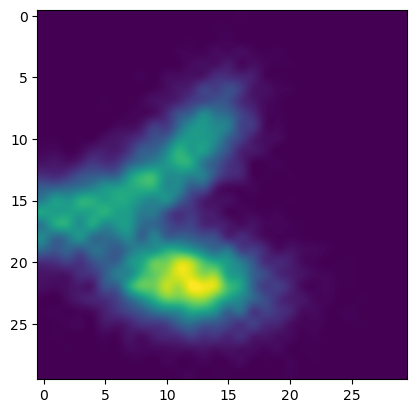}} 
    \subfigure[Diffusion data]{\includegraphics[width=0.233\textwidth]{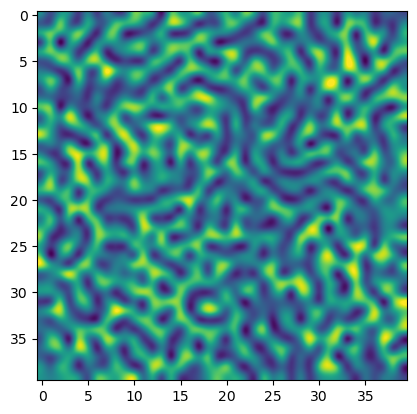}} 
    \subfigure[Coral reef data]{\includegraphics[width=0.233\textwidth]{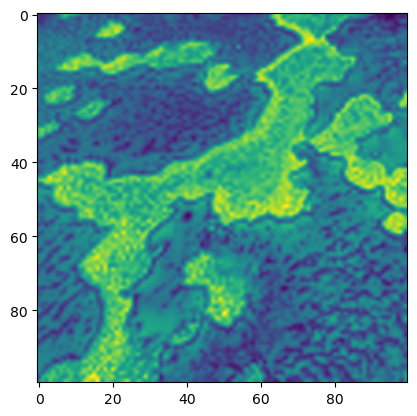}} 
    \caption{Test datasets we used in evaluation, (a) real bathymetry data from Barbados, (b) synthetic cshape distribution, (c) reaction diffusion simulation~\cite{dai2014nonlinear}, and (d) a satellite image of a coral reef.}\vspace{-2.0em}
    \label{fig:test_dataset}
\end{figure}

\textbf{Testing Datasets:} For our testing phase, we consider a set of diverse environments generated both by real-world data and synthetic simulations, as illustrated in Fig.~\ref{fig:test_dataset}. 

\vspace{-1.0em}

\textbf{Performance Metrics:} Many applications in the domain of robotic sampling have a need to sample in information-rich locations at the earliest due to a limited endurance of the robots to carry out the survey, and the environmental processes being sampled are quasi-static. Hence, we measure the \emph{discounted accumulated reward} as one of the performance metrics. 
\emph{Standard deviation of discounted accumulated reward} provides a measure on task sharing between the robots. A lower value implies that the robots contribute more equally to the overall task. 
\emph{Average pair-wise overlap} measures the total pair-wise overlap between the robot trajectories averaged by the number of all possible neighbors. It reflects the performance of a multi-robot sampling approach in terms of coordination and task allocation. 
\emph{Average overlap percentage} measures the percentage of the averaged pair-wise overlap in the robot trajectories and is computed as the ratio between the average pair-wise overlap and the number of steps the robots has traveled. \emph{Coverage} measures the completeness of the active sampling mission, calculated as the ratio between the accumulated reward and the maximum total rewards possible in the given time horizon $H$. \emph{Average communication volume} is a measure of the amount of communication between robots, given by $\frac{1}{N}\sum_{i}\sum_{t}\|\mathcal{N}^{i}_{t}\|,$
where $\mathcal{N}^{i}_{t}$ is the set of neighboring robots of robot $r^{i}$ at time $t$.


\begin{figure}[ht]
    \centering
    \subfigure[Robots trajectories]{\includegraphics[width=0.24\textwidth]{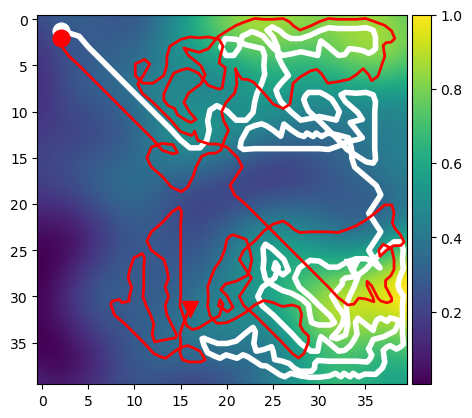}} 
    \subfigure[Discounted accumulated reward]{\includegraphics[width=0.374\textwidth]{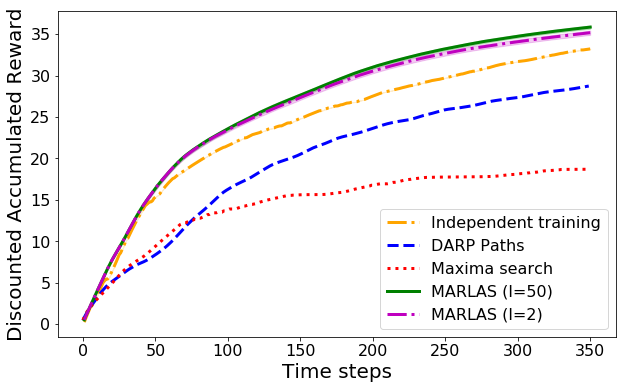}\label{subfig:comparison_dar}} 
    \subfigure[Average pair-wise overlap]{\includegraphics[width=0.374\textwidth]{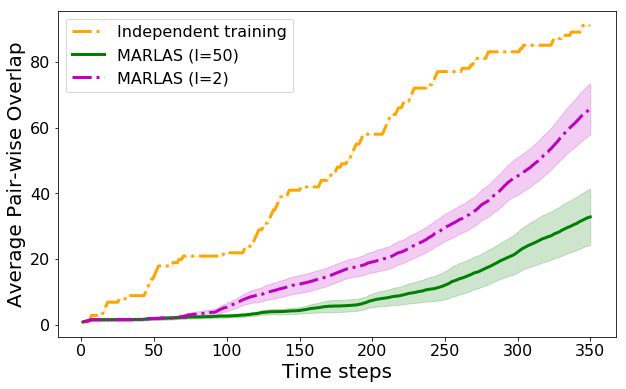}} 
    \subfigure{\includegraphics[width=0.24\textwidth]{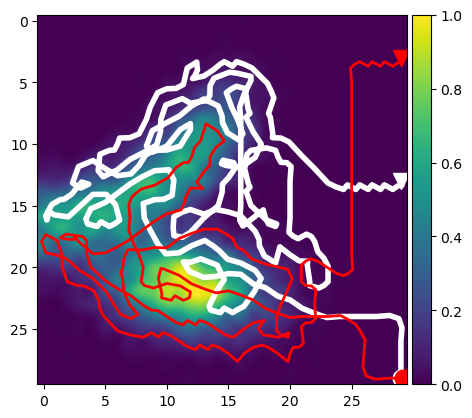}} 
    \subfigure{\includegraphics[width=0.374\textwidth]{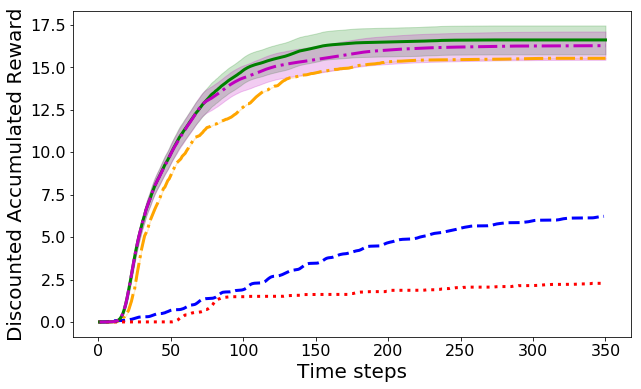}} 
    \subfigure{\includegraphics[width=0.374\textwidth]{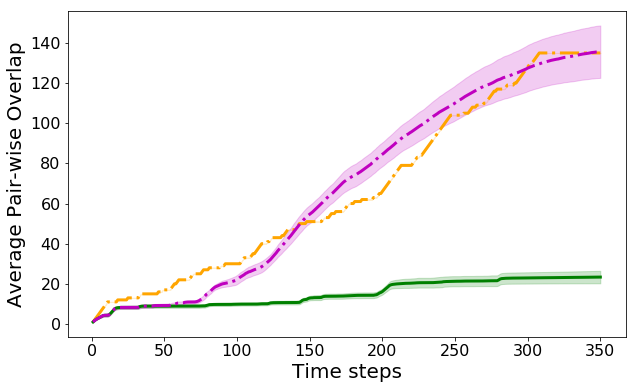}} 
    \subfigure{\includegraphics[width=0.24\textwidth]{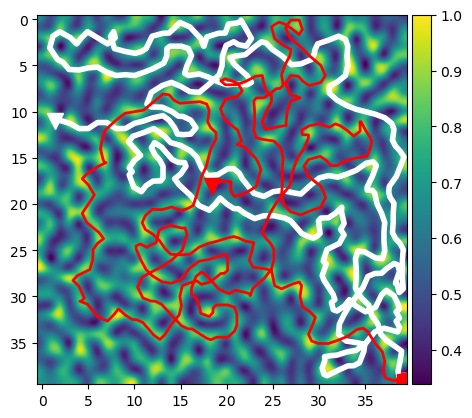}} 
    \subfigure{\includegraphics[width=0.374\textwidth]{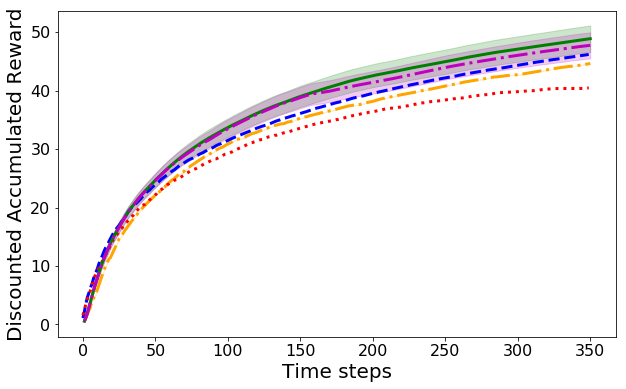}}
    \subfigure{\includegraphics[width=0.374\textwidth]{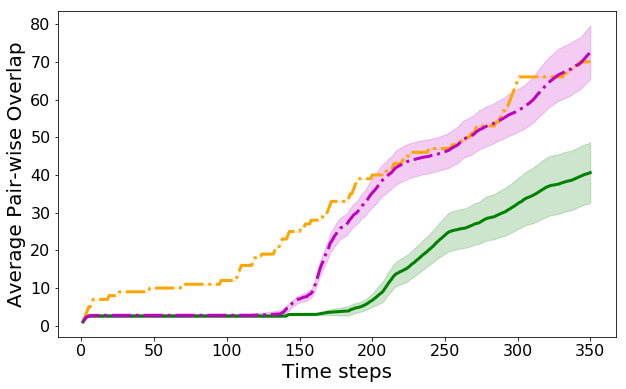}} 
    \caption{Comparison with baseline methods: Column (a) illustrates trajectories of the deployed robot team. Circles are the deployment location, and triangles are the stop location. Column (b) presents comparisons of discounted accumulated reward. Column (c) presents comparisons of average pair-wise overlap. The $95\%$ confidence intervals are over $40$ trials.}\vspace{-2em}
    \label{fig:MetricExps}
\end{figure}

In Fig.~\ref{fig:MetricExps}, we compare MARLAS with multiple baseline sampling methods including independently trained multi-robot sampler~\cite{manjanna2021scalable}, DARP (Divide Area based on the Robot’s initial Positions) algorithm~\cite{kapoutsis2017darp}, and maxima search algorithm~\cite{meghjani2016multi}. We used discounted accumulated reward and average pair-wise overlap metrics for these comparisons. Column (a) of Fig.~\ref{fig:MetricExps} presents the trajectories followed by two robots sampling the underlying distribution using MARLAS. 
The communication radius is fixed as $d_{cr} = 20\% D_{max}$, where $D_{max}$ is the maximum possible distance in the given region of interest.


Columns (b) in Fig.~\ref{fig:MetricExps} presents quantitative comparison of the discounted accumulated rewards. It can be observed that MARLAS 
outperforms baseline methods. 
Fig.~\ref{fig:MetricExps} column (c) compares the pair-wise overlap between paths generated by MARLAS and independently trained multi-robot sampling technique, which keeps track of the full history for inter-robot communications. In these experiments, the robots using MARLAS only communicate history length $l=50$ trajectory data, compared to the baseline methods where the full trajectory history is being sent, MARLAS outperforms other methods in generating non-overlapping utility maximizing paths. Even though the average overlaps for MARLAS with $l=2$ is high, the neighbor position estimate embedded learning helps it to achieve higher discounted rewards compared to the baselines. MARLAS is stochastic because the robots are initialized in the same grid and select random actions to avoid collision. Therefore, its metrics have a tight confidence interval. 

\subsection{Analyzing Scalability of Learned Policy}
\label{subsec:scaling_exps}

Scalability is an important property for a learning-based multi-robot decentralized policy. 
To evaluate the scalability of MARLAS, we conducted experiments with different robot team sizes; different sizes of the workspace; and varying sensing and communication radii. We trained a policy with $5$ robots ($N=5$) and sensing and communication radius $d_{cr}=10$ grids, and generalized to all the different settings.

\begin{figure}
    \centering
    \subfigure[Team Size = $2$]{\includegraphics[width=0.26\textwidth]{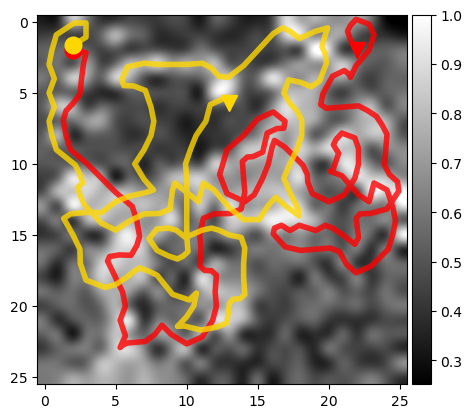}} 
    \subfigure[Team Size = $5$]{\includegraphics[width=0.233\textwidth]{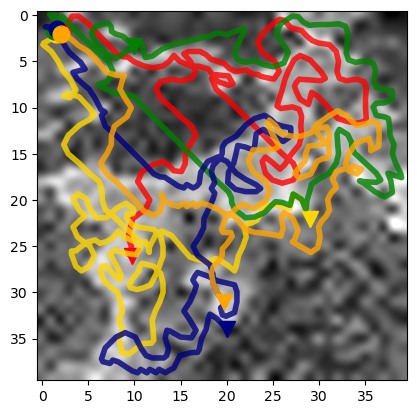}} 
    \subfigure[Team Size = $10$]{\includegraphics[width=0.233\textwidth]{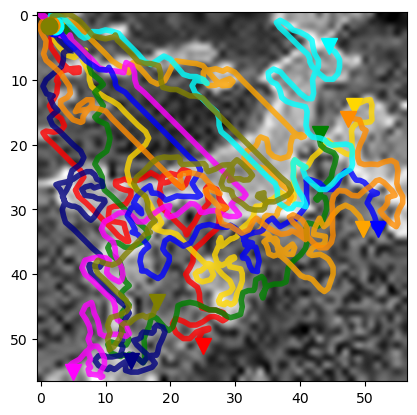}} 
    \subfigure[Team Size = $15$]{\includegraphics[width=0.233\textwidth]{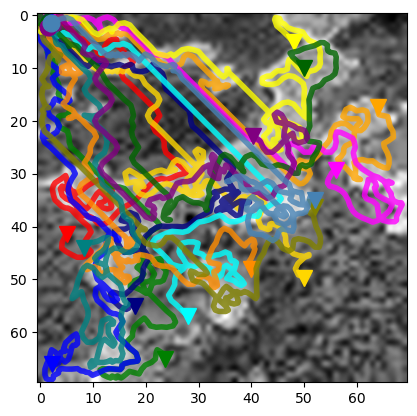}} 
    \subfigure[Time Steps = $10$]{\includegraphics[width=0.233\textwidth]{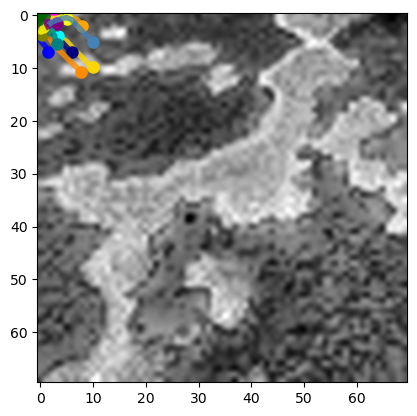}} 
    \subfigure[Time Steps = $40$]{\includegraphics[width=0.233\textwidth]{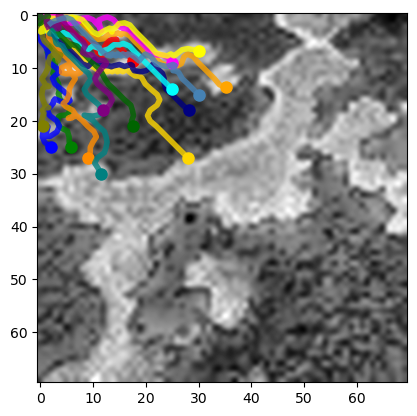}} 
    \subfigure[Time Steps = $80$]{\includegraphics[width=0.233\textwidth]{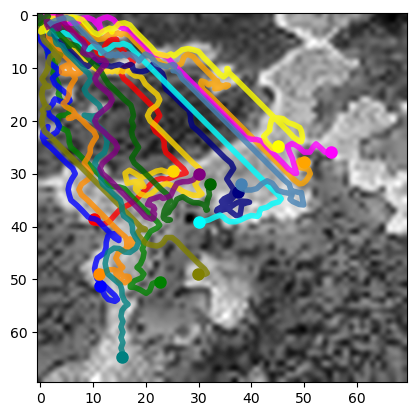}} 
    \subfigure[Time Steps = $150$]{\includegraphics[width=0.233\textwidth]{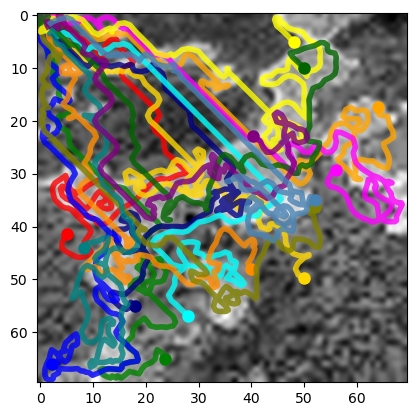}} 
    \caption{Deployment of robot teams over reef data for $150$ time duration. The first row presents robot trajectories (in different colors) as the team size is increased from $2$ to $15$. Circles are the deployment location, and triangles are the stop location. Second row presents sequential progress of a sampling run with $15$ robots at time steps $10, 40, 80$ and $150$. Circles represent the robot current positions.}\vspace{-2.5em}
    \label{fig:num_scaling_figure}
\end{figure}

\textbf{Scalability with the team size:}
\label{subsubsec:scale_size}
In this scenario, we increase the number of robots from $2$ to $20$ and fix the sensing and communication radius to $10$ grids. We scale the map by a factor of $\sqrt{n}$ to maintain a constant density of robots for different sizes of robot teams. For example, a map of size $26\times 26$ is used for a $2$ robot experiment, while a map of size $40\times 40$ is used for a $5$ robot experiment. 
Robots start exploring from the upper left corner of the map, as illustrated in Fig.~\ref{fig:num_scaling_figure} so that they have the same opportunity to accumulate rewards. Qualitative results presented in the first row of Fig.~\ref{fig:num_scaling_figure} show the robot trajectories as the team size is increased from $2$ to $15$. 
The plots in the second row of Fig.~\ref{fig:num_scaling_figure} illustrate the sequential progress of a sampling run with $15$ robots at time steps $10$, $40$, $80$ and $150$. We observe that the robots quickly spread out, identify, and cover to dense reward regions. Furthermore, the robots spread to the different regions and are able to perform the sampling task in a coordinated manner.



\begin{figure}
    \centering
    \subfigure[Standard deviation of discounted reward vs. number of robots]{\includegraphics[width=0.4\textwidth]{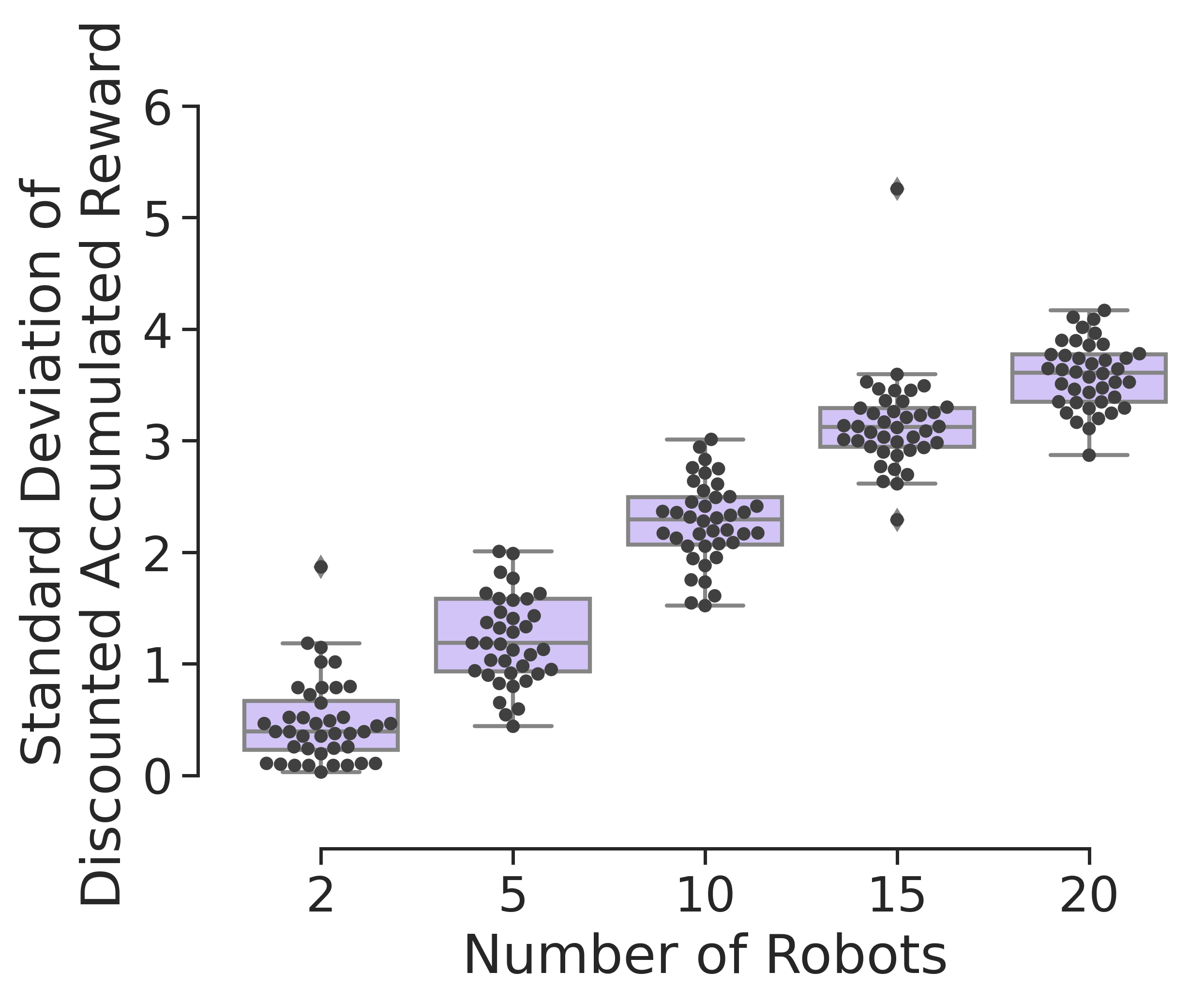}} 
    \subfigure[Coverage vs. number of robots]{\includegraphics[width=0.4\textwidth]{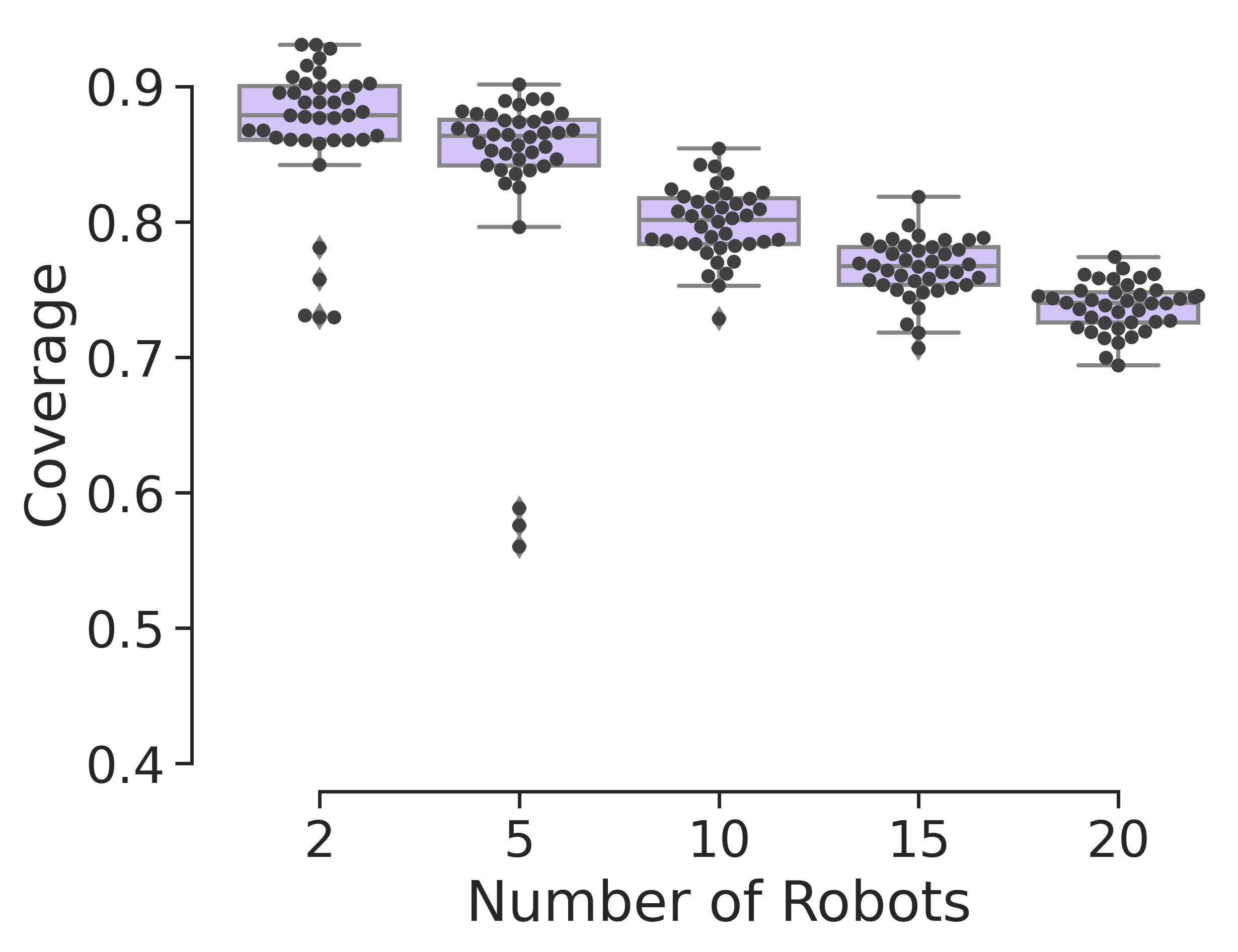}} 
    \caption{Quantitative results for scaling the team size. The dots represent each of the $40$ trials.}\vspace{-2.0em}
    \label{fig:num_scaling_metric}
\end{figure}

The quantitative results in Fig.~\ref{fig:num_scaling_metric} illustrate that the standard deviation of the discounted accumulated rewards start close to $0$ and increase sub-linearly as the robot team size increases. The increase of the standard deviation values is expected as the discounted accumulated rewards also increase as team size increases. We further notice that all the standard deviation values are negligible compared to the discounted accumulated rewards. This indicates that the collected rewards are distributed evenly to each individual robot. Together with the qualitative results, we conclude that each robot continues to contribute equally to the sampling task as we scale up the team size. 
Majority of coverage metrics in Fig.~\ref{fig:num_scaling_metric}(b) remain above $70\%$ regardless of scaling up the number of robots to $20$ and on an $80\times 80$ map, which is far bigger than the training setup. The decrease in coverage is not surprising, as the robots need to travel longer distances between the dense reward regions as the map resolution increases.

\begin{figure}
    \centering
    \subfigure[$d_{cr}=0\% D_{max}$]{\includegraphics[width=0.265\textwidth]{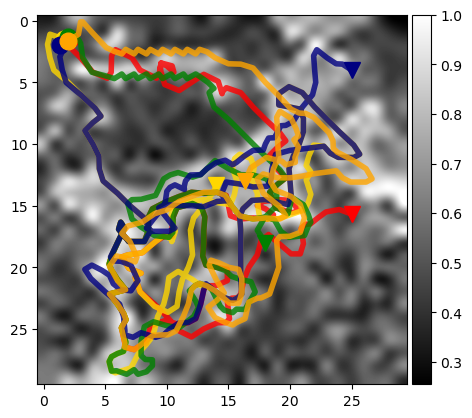}} 
    \subfigure[$d_{cr}=100\% D_{max}$]{\includegraphics[width=0.235\textwidth]{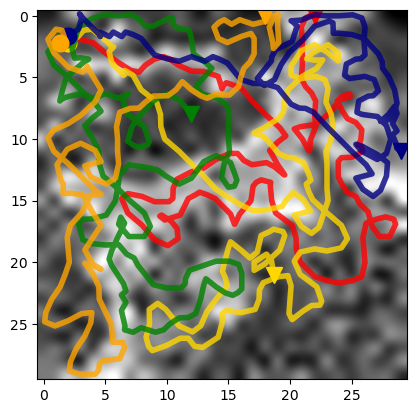}} 
    \subfigure[Metrics v.s. Communication radii]{\includegraphics[width=0.48\textwidth]{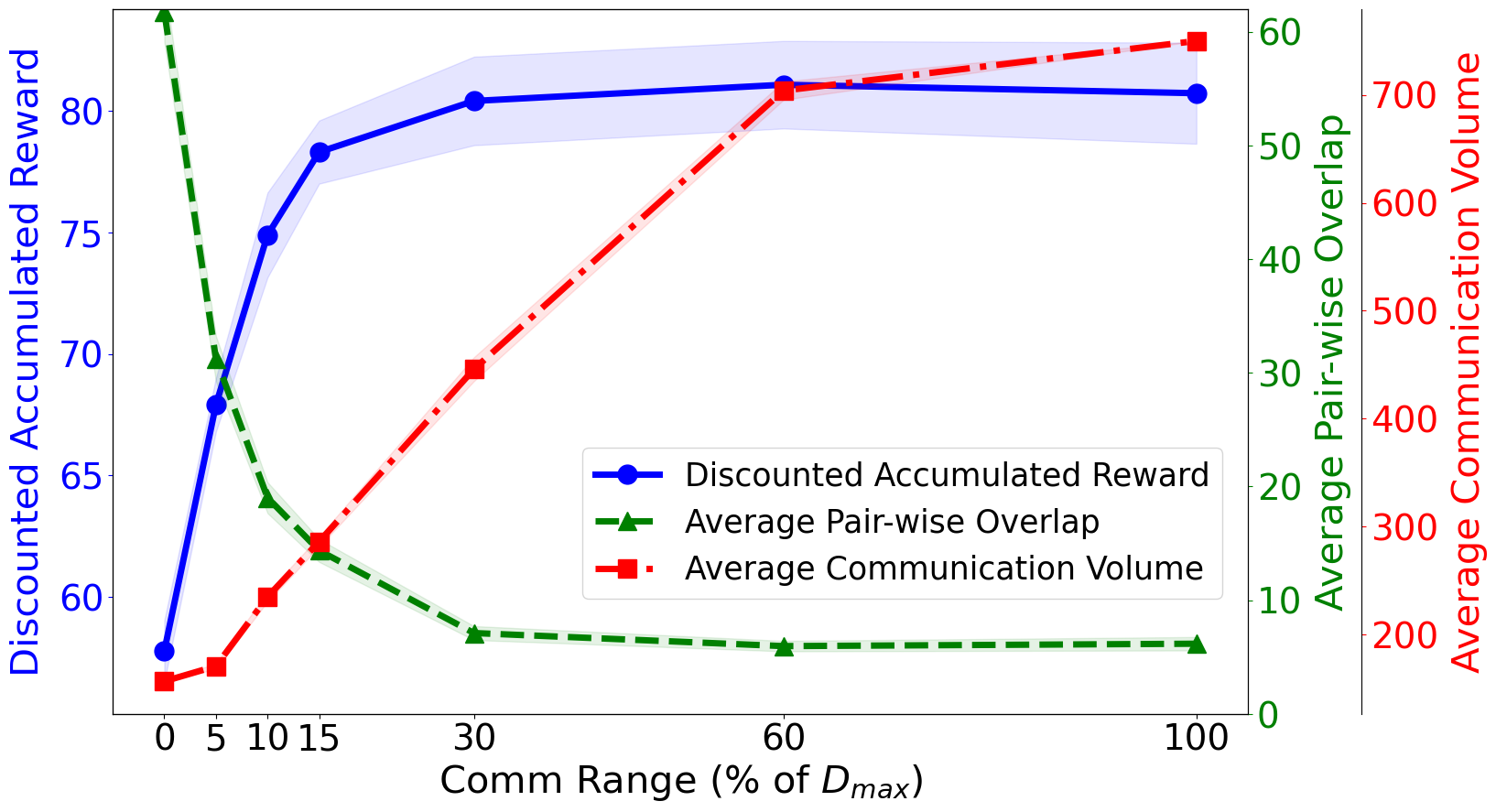}} 
    \caption{(a) and (b) Robot team with $5$ robots carrying out the sampling task with $d_{cr}=0\% D_{max}$ and $d_{cr}=100\% D_{max}$ using the learned policy. Circles are the deployment location, and triangles are the stop location. (c) Quantitative results include plots of discounted accumulated reward, average pair-wise overlap, and average communication volume over different communication radii. The $95\%$ confidence intervals over $40$ trials.}\vspace{-1.5em}
    \label{fig:comm_scaling_metric}
\end{figure}

\textbf{Scalability with sensing and communication radius:}
\label{subsubsec:scale_comm}
To evaluate the scalability with the sensing and the communication radius, we change the sensing and the communication radius $d_{cr}$ during the deployment from $0\% D_{max}$ to $100\% D_{max}$. In our simulations, we down-sampled the reef data to $30\times 30$. Fig.~\ref{fig:comm_scaling_metric} (a) and (b) present $5$ robots deployed at upper left corner of the map carrying out the sampling task with $d_{cr}=0\% D_{max}$ and $100\% D_{max}$ respectively. As expected, with a larger sensing and communication radius, robots explore cooperatively and do not sample the same region repeatedly.

The plots in Fig.~\ref{fig:comm_scaling_metric}(c) illustrate that the discounted accumulated reward increases and the average pair-wise overlap decreases, and both quickly plateau at $d_{cr}=30\% D_{max}$. The average communication volume increases linearly and slows its rate after $d_{cr}=60\% D_{max}$. We observed consistent behaviors over all the other test datasets. We conclude that, with a moderate range of sensing and communication, the cooperative behavior is achieved by the team, and increasing the sensing and communication radius yields no significant contribution to its performance.

\subsection{Analyzing Robustness of the Learned Policy}
\label{subsec:robustness_exps}
In this section, we show how the ability to estimate the position of other robots makes the system robust against communication failures and robot failures. 

\begin{figure}
    \centering
    \subfigure[Discounted accumulated reward for different robot failure conditions]{\includegraphics[width=0.3\textwidth]{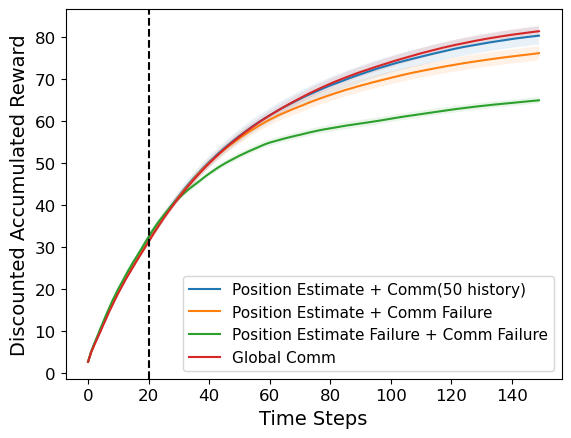}} 
    \subfigure[Average overlap percentage for different robot failure conditions]{\includegraphics[width=0.3\textwidth]{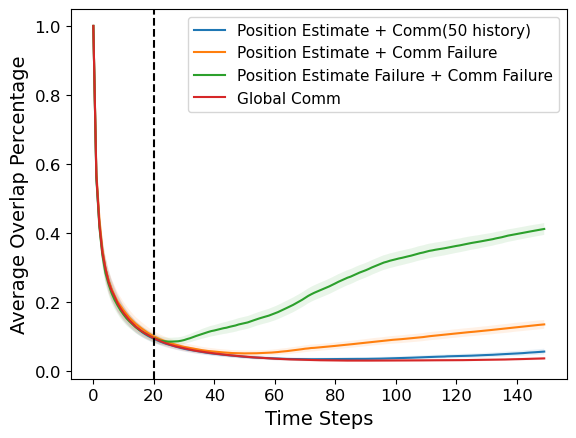}} 
    \subfigure[Robot failure experiments]{\includegraphics[width=0.38\textwidth]{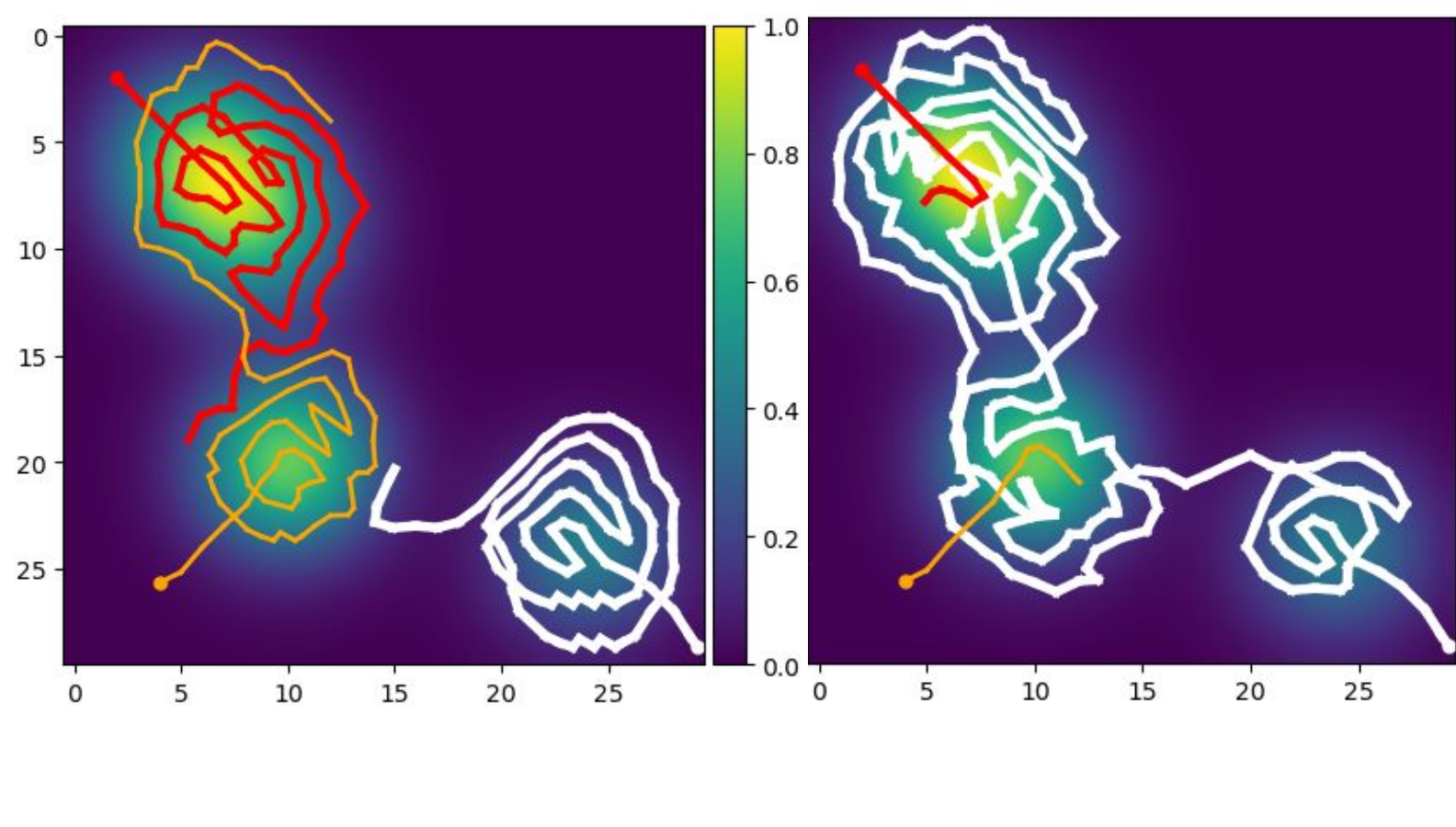}} 
    \caption{Robustness to failures: The failures happen at time step $20$, denoted as the dashed line. The $95\%$ confidence intervals are computed over $40$ trials. (c) Snapshots of robot failure experiments. 
    }\vspace{-1.5em}
    \label{fig:comm_robust_metric}
\end{figure}

\textbf{Communication Failure:}
\label{subsubsec:robust_comm}
We first consider a communication failure scenario where the robot can only rely on sensing to localize neighbor positions. The robot team and workspace setup is kept same as the one described in the communication scalability experiment. We created four scenarios by changing the neighbor position estimation and communication capability of the learned model: the first is when the estimates of the neighbor positions and the communication are both enabled; The second scenario is when the communication failed at the time step $20$ and robots localize their neighbors only via filtering. The third scenario is when both the estimates of the neighbor positions and the communication failed at time step $20$. Lastly, we use the control policy with global communication as a benchmark. To better reflect the overlap change with respect to the robots' travel distance, we use average overlap percentage as a metric. The results in Fig.~\ref{fig:comm_robust_metric} (a) and (b) illustrate that before the failure, there is no significant difference between models' performances in all the scenarios. Once the failure is introduced, the second scenario, with the communication failure but active position estimate, has a performance that is very close to that of the benchmark scenario with the global communication. This means that the estimates of the neighbor positions serve as good substitutes for the ground truth information. The scenario with both the communication failure and the neighbor position estimation failure performs poorly compared to all the other scenarios showing that the proposed belief model compensates for complete communication failures, thus increases the robustness of MARLAS. We observed consistent results over all the test datasets.



\textbf{Robot Failure:}
\label{subsubsec:robot_failure}
In Fig.~\ref{fig:comm_robust_metric} (c), we present qualitative experimental results to show robustness of MARLAS against the robot failure. As shown in left image in Fig.~\ref{fig:comm_robust_metric} (c), when there are no failures, all three robots share and cover the hotspot regions. When two of the robots fail during a survey, the other robot samples from the hotspot regions and achieves good overall performance for the team as depicted in the right image in Fig.~\ref{fig:comm_robust_metric} (c). These preliminary results display a fault-tolerance nature of MARLAS and we would like to further investigate this behavior in the near future.



\begin{figure}
    \centering
    \subfigure[$t=0$]{\includegraphics[width=0.298\textwidth]{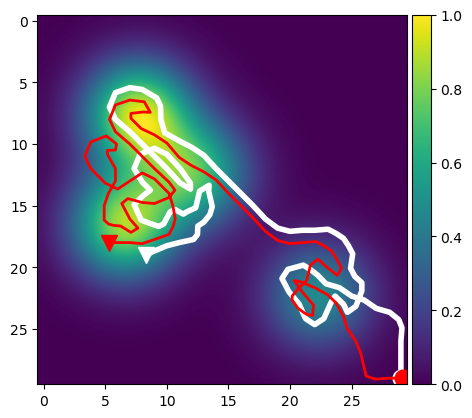}} 
    \subfigure[$t=100$]{\includegraphics[width=0.264\textwidth]{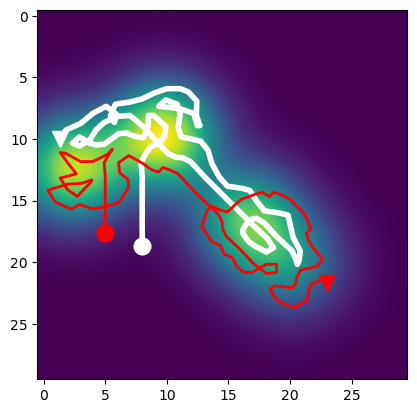}} 
    \subfigure[$t=200$]{\includegraphics[width=0.264\textwidth]{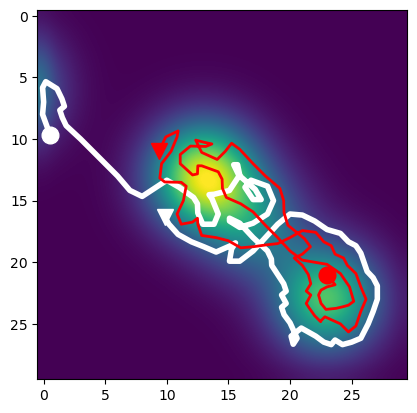}} 
    \caption{Online adaptation: Three sequential time steps of a changing field and MARLAS generating robot paths that adaptively cover the underlying distribution. Circles represent the start locations, and triangles represent the stop locations.}\vspace{-2em}
    \label{fig:online_adaptation}
\end{figure}
\subsection{Online Adaptations to the changing field}
To demonstrate the online adaptation capability of MARLAS, we deploy the robots to sample a stochastic field. We synthesize a dynamic field as a mixture-of-Gaussians with a random motion added to the means of the Gaussians. Fig.~\ref{fig:online_adaptation} illustrates three sequential time steps of a changing field and MARLAS generating robot paths that adaptively cover the underlying distribution. An updated low-resolution prior is provided to the robots every $100$ time steps which can be achieved by either an aerial vehicle or a sparse network of stationary sensors. The MARLAS algorithm is able to generate updated plannings for the robot team on the fly. 

\subsection{Robot Experiments}
\begin{figure}
    \centering
    \includegraphics[width=0.85\textwidth]{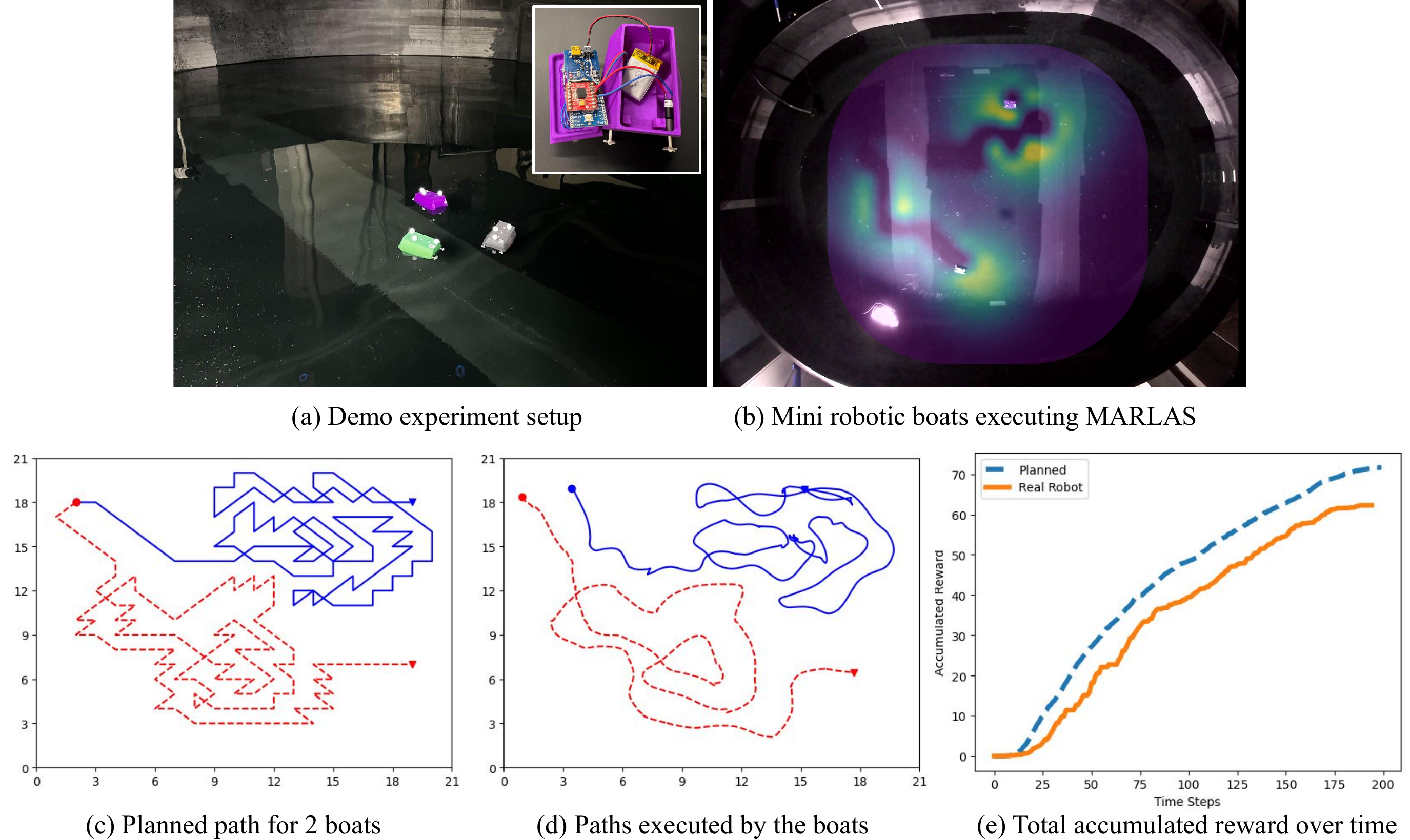}
    \caption{ (a) Indoor water tank experimental setup and embeded picture shows the internals of a mini robotic boat, (b) MARLAS policies run on mini robotic boats to sample from the spatial field that is projected over the water surface.}
    \vspace{-2em}
    \label{fig:demo}
\end{figure}

To demonstrate the feasibility of the proposed strategy in the physical world, we conducted experiments in an indoor water tank using centimeter scale robotic boats as presented in Fig.~\ref{fig:demo}(a) and (b). We pre-plan the robot trajectories using the MARLAS algorithm, and each mini boat is controlled by base station commands to execute its assigned trajectory. Plots in Fig.~\ref{fig:demo}(c), (d), and (e) illustrate the difference between planned path and the path executed by the physical robots. These results indicate that the action space used for our policy generation needs further investigation to include the dynamics of the autonomous vehicles. In future, we plan to design the policies over motion primitives to achieve smooth trajectories for physical robots.

\section{Conclusion and Future Work}

We proposed an online and fully decentralized multi-robot sampling algorithm that is robust to communication failures and robot failures, and is scalable with both the size of the robot team and the workspace size. Our proposed method, MARLAS, outperforms other baseline multi-robot sampling techniques. We analyze the key features of MARLAS algorithm and present qualitative and quantitative results to support our claims. In the future, we like to investigate the application of MARLAS to sample dynamic spatiotemporal processes. We would like to study the effect of reducing the length $l$ of trajectory history that needs to be communicated between the teammates and thus enhance the memory efficiency and system robustness. We would like to further investigate distributed learning with map exploration using local measurements of the environment. 

\section*{Acknowledgements}
We gratefully acknowledge the support of NSF IIS 1812319 and ARL DCIST CRA W911NF-17-2-0181.


%
%
%





%

%

%
%

\bibliographystyle{IEEEtran} 
\bibliography{refs} 

\end{document}